\documentclass[10pt,twocolumn,letterpaper]{article}

\usepackage[dvipsnames]{xcolor}
\usepackage{btas}
\usepackage{times}
\usepackage{epsfig}
\usepackage{graphicx}
\usepackage{amsmath}
\usepackage{amssymb,adjustbox,ctable,threeparttable,booktabs,mathrsfs,verbatim,algorithm,algpseudocode}

\usepackage[pagebackref=true,breaklinks=true,letterpaper=true,colorlinks,bookmarks=false]{hyperref}

\btasfinalcopy 

\ifbtasfinal\fi
\begin{document}

\title{Automatic Latent Fingerprint Segmentation}

\author{Dinh-Luan Nguyen, Kai Cao and Anil K. Jain\\
Michigan State University\\
East Lansing, Michigan, USA\\
{\tt\small nguye590@msu.edu, \{kaicao, jain\}@cse.msu.edu}
}

\maketitle
\thispagestyle{empty}

\begin{figure}[b]
\parbox{\hsize}{\em
978-1-5386-7180-1/18/\$31.00 \ \copyright 2018 IEEE.
}\end{figure}

\begin{abstract}
   We present a simple but effective method for automatic latent fingerprint segmentation, called SegFinNet. SegFinNet takes a latent image as an input and outputs a binary mask highlighting the friction ridge pattern. Our algorithm combines fully convolutional neural network and detection-based approaches to process the entire input latent image in one shot instead of using latent patches. Experimental results on three different latent databases (i.e. NIST SD27, WVU, and an operational forensic database) show that SegFinNet outperforms both human markup for latents and the state-of-the-art latent segmentation algorithms. We further show that this improved cropping boosts the hit rate of a latent fingerprint matcher.
\end{abstract}

\section{Introduction}

Latent fingerprints, also known as fingermarks, are friction ridge impressions formed as a result of someone touching a surface, particularly at a crime scene. Latents have been successfully used to identify suspects in criminal investigations for over 100 years by comparing the similarity between latent and rolled fingerprints in a reference database~\cite{jain201650}. Latent cropping (segmentation) is the crucial first step in the latent recognition algorithm. For a given set of latent enhancement, minutiae extraction, and matching modules, different cropping masks for friction ridges can lead to dramatically different recognition accuracies. Unlike rolled/slap fingerprints, which are captured in a controlled setting, latent fingerprints are typically noisy, distorted and have low ridge clarity. This creates challenges for an accurate automatic latent cropping algorithm.

\begin{figure}[!tp]
\centering
\includegraphics[width=8cm]{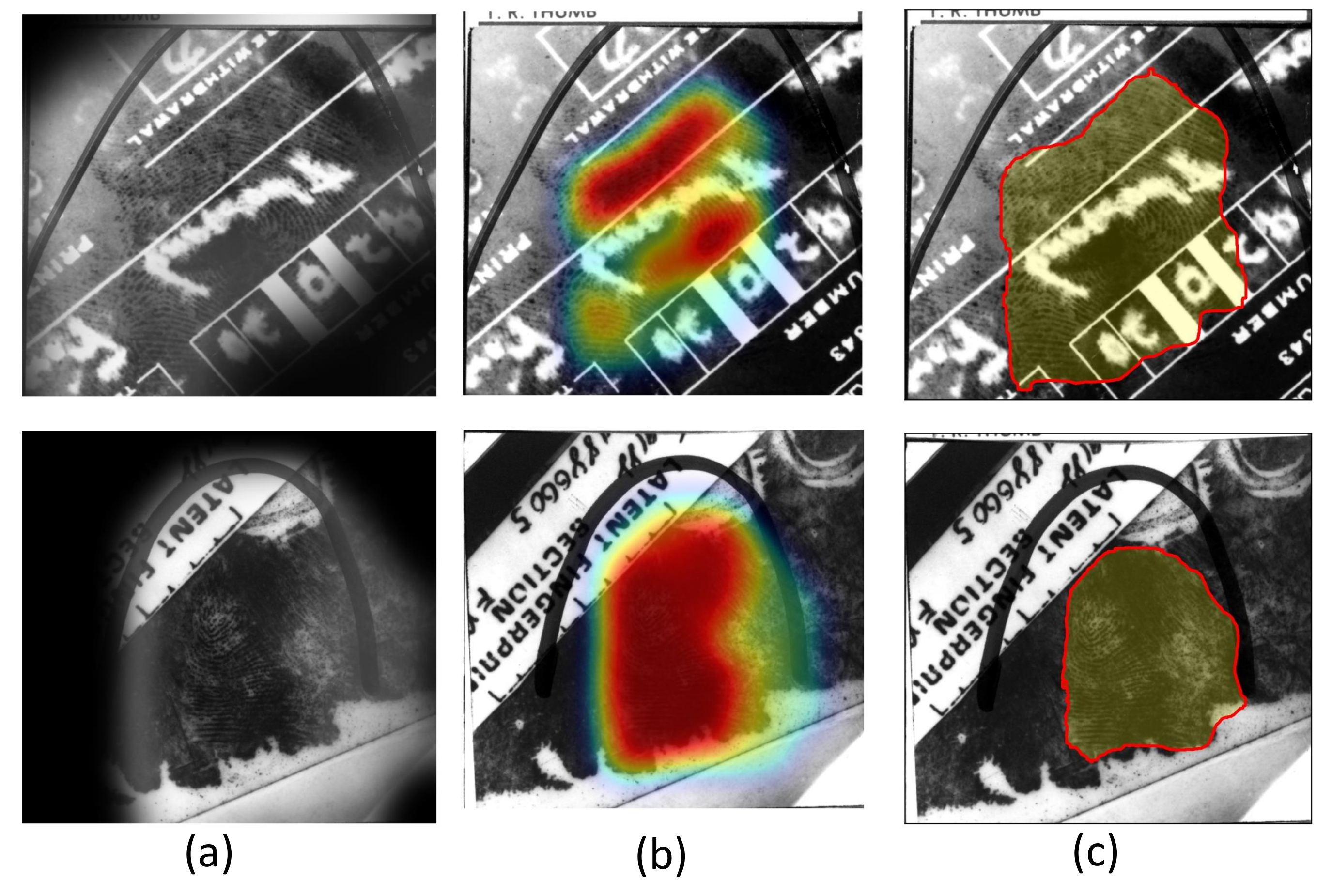}
\caption{SegFinNet with visual attention mechanism for two different input latents, one per row: (a) Focused region from Visual attention module (section \ref{sec:wheretolook}); (b) Original latents overlaid with a heat map showing the probability of occurrence of friction ridges (from \textcolor{red}{high} to \textcolor{blue}{low}); (c) Binary mask (boundary marked in red) used in subsequent modules: enhancement, feature extraction, and matching.}
\vspace{-1em}
\label{fig:Vis_intro}
\end{figure}

We map the latent fingerprint cropping problem to a sequence of computer vision tasks as follow: (a) \emph{Object detection}~\cite{ren2017faster} as friction ridge localization; (b) \emph{Semantic segmentation}~\cite{long2015fully} as separating all possible friction ridge patterns (foreground) from the background; and (c) \emph{Instance Segmentation}~\cite{he2017mask} as separating individual friction ridge patterns in the input latent by semantic segmentation.

Object segmentation can be based on two different approaches: (i) fully convolutional neural networks (FCN) based~\cite{long2015fully} and (ii) object detection based~\cite{he2017mask}. FCN based segmentation consists of a series of consecutive receptive fields in its network and is built on translation invariance. Instead of computing general nonlinear functions, FCN builds its nonlinear filters based on relative spatial information in a sequence of layers. On the other hand, detection based segmentation first finds a core and then branches out in parallel to construct pixel-wise segmentation from regions of interest returned by previous detection.

\begin{table*}[!tp]
	\centering   
	\label{tab:existing_approach}
	\begin{footnotesize}
		\begin{threeparttable}
	\begin{adjustbox}{max width=\textwidth}
		\begin{tabular}{|m{1.6cm}|m{2.6cm}|m{2.9cm}|m{4.8cm}|m{3.5cm}|}
			\hline
			{\bfseries Study}& {\bfseries Method}& {\bfseries Database}& {\bfseries Results}& {\bfseries Comments}\\
			\hline\hline
			
			{Choi \etal~\cite{choi2012automatic}}&Patch orientation and ridge frequency&NIST SD27 and WVU;\newline Background: 32K images&NIST SD27: 14.78\% MDR; 47.99\% FDR \tnote{(+)}\newline WVU: 40.88\% MDR; 5.63\% FDR \newline \emph{Matching:} 16.28\% on NIST SD27 and 35.1\% on WVU with COTS tenprint matcher \tnote{(*)}&Relies on input image quality and orientation estimation\\
			
			\hline
			{Zhang \etal~\cite{zhang2013adaptive}}&Adaptive directional total variance model&NIST SD27 (1,000 dpi);\newline Background: 27K images&14.10\% MDR; 26.13\% FDR;\newline \emph{Matching:} 2\% on NIST SD27 with Verifinger SDK 6.6&Relies on orientation field and orientation coherence estimation\\
			
			\hline			
			{Ruangsakul \etal~\cite{ruangsakul2015latent}}&Fourier Subbands using spatial-frequency information&NIST SD27;\newline Background: 27K images&31.90\% MDR; 32.50\% FDR;\newline \emph{Matching:} 14\% on NIST SD27 with Verifinger SDK 6.6&Handcrafted subband features; dilation and erosion used to fill gaps and eliminate islands\\
			
			\hline
			{Cao \etal~\cite{cao2014segmentation}}&Patch classification based on learned dictionary&NIST SD27 and WVU;\newline Background: 32K images&\emph{Matching:} 61.24\% on NIST SD27 and 70.16\% on WVU with a COTS matcher \tnote{(*)}&Heuristic patch classification; relies on learned dictionary quality and convex hull to get smooth mask\\
			
			\hline			
			{Liu \etal~\cite{liu2016latent}}&Linear density on a set of line segments from the texture component of latent images&NIST SD27;\newline Background: 27K images&13.32\% MDR; 24.21\% FDR;\newline \emph{Matching:} 22\% on NIST SD27 with Verifinger SDK 4.3&Use dilation and erosion for post-processing and use convex hull to get smooth mask\\
			
			\hline			
			{Zhu \etal~\cite{zhu2017latent}}&Neural network as binary patch based classifier&NIST SD27;\newline No background reported&10.94\% MDR; 11.68\% FDR;\newline No matching accuracy reported&Relies on neural network classifier; patch by patch processing is time consuming\\
			
			\hline			
			{Ezeobiejesi \etal~\cite{ezeobiejesi2017latent}}&Patch-based stack of restricted Boltzmann machines&NIST SD27, WVU, and IIITD;\newline No background reported&NIST SD27: 1.25\% MDR; 0.04\% FDR \tnote{(\#)};\newline WVU: 1.64\% MDR; 0.60\% FDR;\newline IIITD: 1.35\% MDR; 0.54\% FDR;\newline No matching accuracy reported&Depends on the stability of classifier; time consuming\\
			
			\specialrule{.15em}{.1em}{.1em}
			{Proposed \newline approach}&Automatic segmentation based on FCN and detection based fusion&NIST SD27, WVU, and a forensic database;\newline Background: 100K images&MDR, FDR, and IoU metrics;\newline \emph{Matching:} 70.8\% on NIST SD27 and 71.3\% on WVU with a COTS matcher;\newline \emph{Matching:} 12.6\% on NIST SD27 and 28.9\% on WVU with Verifinger SDK 6.3 on 27K images&Non-patch based approach; non-warp region of interest; visual attention mechanism; voting masks technique\\
			\specialrule{.15em}{.1em}{.1em}
		\end{tabular}
		\end{adjustbox}
		\begin{tablenotes}
		\scriptsize
			\item[(+)] \scriptsize{MDR: Missed Detection Rate; FDR: False Detection Rate; IoU: Intersection Over Union}
			\item[(*)] \scriptsize{COTS: Commercial off the shelf; The authors did not identify which COTS was used.}
			\item[(\#)] \scriptsize{This work used a subset of dataset for training and their metrics are defined on patches.}
		\end{tablenotes}
		\end{threeparttable}
	\end{footnotesize}
	\caption{Published works related to latent fingerprint segmentation.}
	\vspace{-1em}
\end{table*}

Our proposed method, called SegFinNet, inherits the idea of instance segmentation and utilizes the advantages of FCN~\cite{long2015fully} and Mask-RCNN~\cite{he2017mask} to deal with the latent fingerprint cropping problem. SegFinNet uses Faster RCNN~\cite{ren2017faster} as its backbone while its head comprises of atrous transposed convolution layers~\cite{chen2016deeplab}. We utilize a combination of a non-warp region of interest technique, a fingerprint attention mechanism, a fusion voting and a feedback scheme to take advantage of both the deep information from neural networks and the shallow appearance of fingerprint domain knowledge (Figure \ref{fig:Vis_intro}). In our experiments, SegFinNet shows a significant improvement not only in latent cropping, but also in latent search (see Section \ref{sec:Matching} for more details).

\begin{figure*}[!tbp]
\centering
\includegraphics[width=16.5cm]{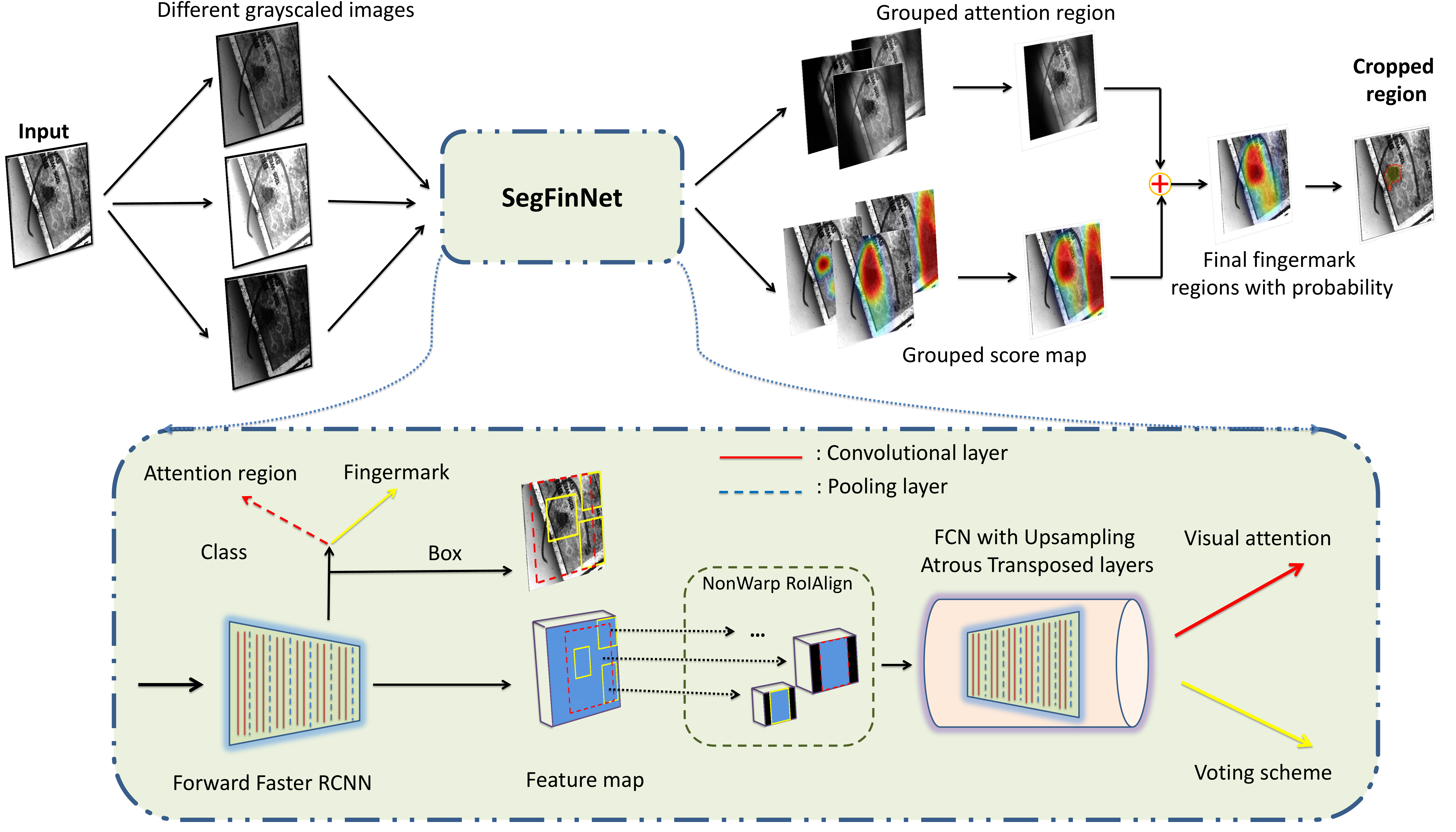}
\caption{SegFinNet architecture.}
\vspace{-1em}
\label{fig:Full_architecture}
\end{figure*}

\section{Related work}
\label{sec:Related_works}
In latent fingerprint recognition literature, it is a common practice to use a patch based approach in various modules, (i.e. minutiae extraction~\cite{Nguyen2018MinutiaeNet} and enhancement~\cite{zhang2013adaptive, cao2014segmentation}). In such cases, the input latent is divided into multiple overlapping patches at different locations. The latent segmentation module of latent recognition systems has also been approached in this patch based manner, both with convolutional neural networks (convnet) and non-convnet approaches. Table \ref{tab:existing_approach} concisely describes these methods reported in the literature. 

\textbf{Non-convnet patch-based approaches:}
Choi \etal~\cite{choi2012automatic} constructed orientation and frequency maps to use as a reference in evaluating latent patches. Essentially, this is a dictionary look up map aimed at classifying each individual patch into two classes. Zhang \etal~\cite{zhang2013adaptive} used an adaptive directional total variance model which also relies on the orientation estimation. From the information in the spatial-frequency domain, Ruangsakul \etal~\cite{ruangsakul2015latent} proposed a Fourier subband method with necessary post-processing to fill gaps and eliminate islands. Cao \etal~\cite{cao2014segmentation} classified patches based on a dictionary which depends on dictionary quality and needs post-processing to make the masks smooth. Liu \etal~\cite{liu2016latent} utilized texture information to develop linear density on a set of line segments but requires a post-processing technique.

The features used in all the above methods are ``hand-crafted'' and rely on post-processing techniques. With the success of deep neural networks in many domains, latent fingerprint cropping has also been tackled using them.

\textbf{Convnet patch-based approaches:}
Zhu \etal~\cite{zhu2017latent} used a classification neural network framework to classify patches. This approach is similar to existing non-convnet methods except it simply replaces hand-crafted features by convnet features. Ezeobiejesi \etal~\cite{ezeobiejesi2017latent} used a stack of restricted Boltzmann machines in an idea similar to ~\cite{zhu2017latent}.

\begin{figure}[!tbp]
\centering
\includegraphics[width=8.5cm]{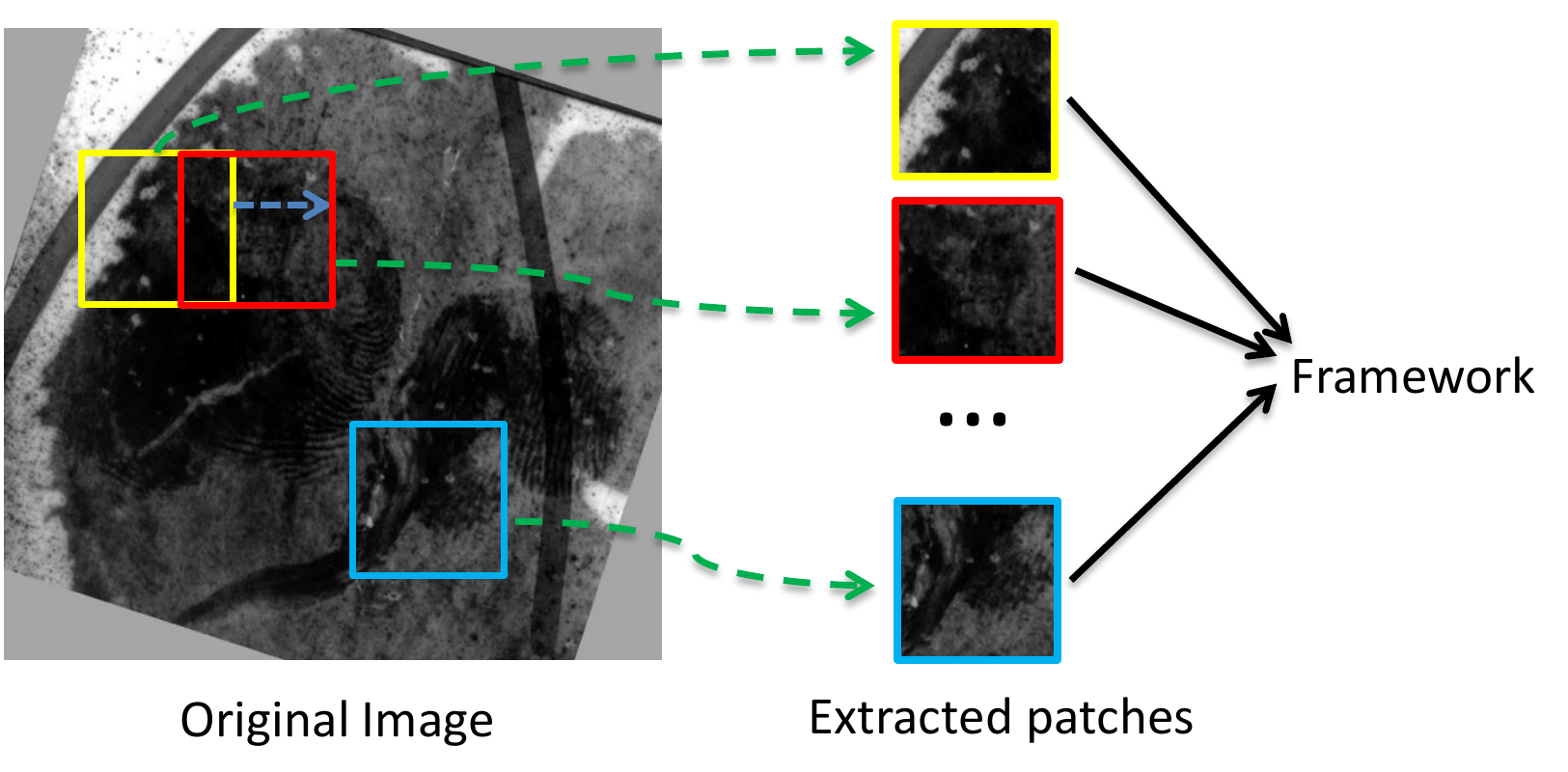}
\caption{General pipeline of patch-based approaches.}
\vspace{-1em}
\label{fig:patchBased}
\end{figure}

There are a number of disadvantages to patch-based approaches. (i) Patch-based methods take significant time to compute since they need to process every patch into the framework (Figure \ref{fig:patchBased}). Typically, a patch size is around $96 \times 96$. Thus, on average there are, approximately $500$ patches in a latent fingerprint in the NIST SD27 dataset. That means patch-based approaches process $500$ subimages instead of one.
(ii) Patch-based approaches cannot separate multiple instances of friction ridge patterns, i.e. more than one latent (overlapping or non-overlapping) in the input image.

Our work combines fully convolutional neural network and detection based approaches for latent fingerprint segmentation to process the entire input latent image in one shot. Furthermore, it also utilizes a top-down approach (detection before segmentation) which can also be applied to segmenting overlapping latent fingerprints. The main contributions of this paper are as follows:
\vspace{-1em}
\begin{itemize}
\item A fully automatic latent segmentation framework, called \emph{SegFinNet}, which processes the entire input image in one shot. It also outputs multiple instances of fingermark locations.
\vspace{-0.5em}
\item \emph{NonWarp-RoIAlign} is proposed to obtain more precise segmentation while mapping region of interest (cropped region) in the feature map to the original image.
\vspace{-0.5em}
\item \emph{Visual attention technique} is designed to focus only on fingermark regions in the input image. This addresses the problem of \emph{``where to look''}.
\vspace{-0.5em}
\item \emph{Feedback scheme} with weighted loss is utilized to emphasize the difference in importance of different objective functions (foreground-background, bounding box, etc.)
\vspace{-0.5em}
\item \emph{Majority voting fusion mask} is proposed to increase the stability of the cropped mask while dealing with different qualities of latents.
\vspace{-0.5em}
\item Experiments demonstrating that the proposed framework outperforms both human latent cropping and published automatic cropping approaches. Furthermore, the proposed segmentation framework, when integrated with a latent AFIS, boosts the search accuracy on three different latent databases: NIST SD27, WVU, and MSP DB (an operational forensic database).
\end{itemize}

\section{SegFinNet}
\label{sec:SegFinNet}
Based on the idea of detection-based segmentation of Mask RCNN~\cite{he2017mask}, we build our framework upon the Faster RCNN architecture~\cite{ren2017faster}, where the head is a series of atrous transposed convolutions for pixel-wise prediction.

Unlike previous patch-based approaches which used either handcrafted features~\cite{ruangsakul2015latent, cao2014segmentation, liu2016latent} or a convnet approach~\cite{zhu2017latent, ezeobiejesi2017latent}, we feed the whole input latent image once to Faster RCNN and process candidates foreground regions returned by SegFinNet. This reduces the training time, and avoids post-processing heuristics to combine results from different patches. Figure \ref{fig:Full_architecture} and Table \ref{Framework_algorithm} illustrate the SegFinNet architecture in details. 

\begin{algorithm}[!tbp]
\caption{SegFinNet latent fingerprint cropping}\label{Framework_algorithm}
\begin{algorithmic}[0]
\State \textbf{Input:} Latent fingerprint image
\State \textbf{Output:} Binary mask
\end{algorithmic}

\begin{algorithmic}[1]
\State Generate different types of grayscale images.
\Procedure{Process each grayscale image}{}
\State Feed the input image to Faster RCNN to obtain the feature map together with the bounding boxes (coordinates) of fingermarks and attention region candidates.
\For{\textbf{each} box in the candidate list}
	\State Regard each box as a friction ridge  image to feed to FCN to obtain Visual attention region (section \ref{sec:wheretolook}) and Voting scheme (section \ref{sec:Voting_fusion_masks}) results.
\EndFor	
\EndProcedure
\State Fuse results to get the final fingermark probabilities.
\State Apply a hard-threshold to get binary mask for input latent image.
\end{algorithmic}
\end{algorithm}

\subsection{NonWarp-RoIAlign}
\label{sec:NonWarp}
The RoIAlign module in Mask RCNN can handle the misalignment problem\footnote{Due to mapping each point in feature map to the nearest value in its neighboring coordinate grid. We refer readers to~\cite{he2017mask} for more details.} while quantizing the region of interest (RoI) coordinates in feature maps by using bilinear interpolation on fixed point values~\cite{jaderberg2015spatial}. However, it warps the RoI feature maps into a square size (e.g. $96 \times 96$) before feeding to the upsampling step. This leads to further misalignment and information loss when reshaping the ROI feature map back to original size in image coordinates.

The idea of NonWarp-RoIAlign is simple but effective. Instead of warping RoI feature maps into squared size and then applying multiple deconvolution (upsampling) layers, we only pad zero value pixels to get to a specific size. This can avoid the loss of pixel-wise information when warping regions. We use atrous convolution~\cite{chen2016deeplab} when upsampling for faster processing and saving memory resources (see Figure \ref{fig:Full_architecture} for more visualization). The advantage of this method of upsampling is that we can deal with the multi-scale problem with atrous spatial pyramid pooling properties and weights of atrous convolution can be obtained from the transposed corresponding forward layer.

We also adopt the strategy of combining high-level layers with low-level layers~\cite{he2016deep, chen2016deeplab, Nguyen2018MinutiaeNet} to get finer detail prediction while maintaining high-level semantic interpretation as multi-scale prediction.

\subsection{Where to look? Visual attention mechanism}
\label{sec:wheretolook}
\begin{figure}[!bp]
\centering
\includegraphics[width=8cm]{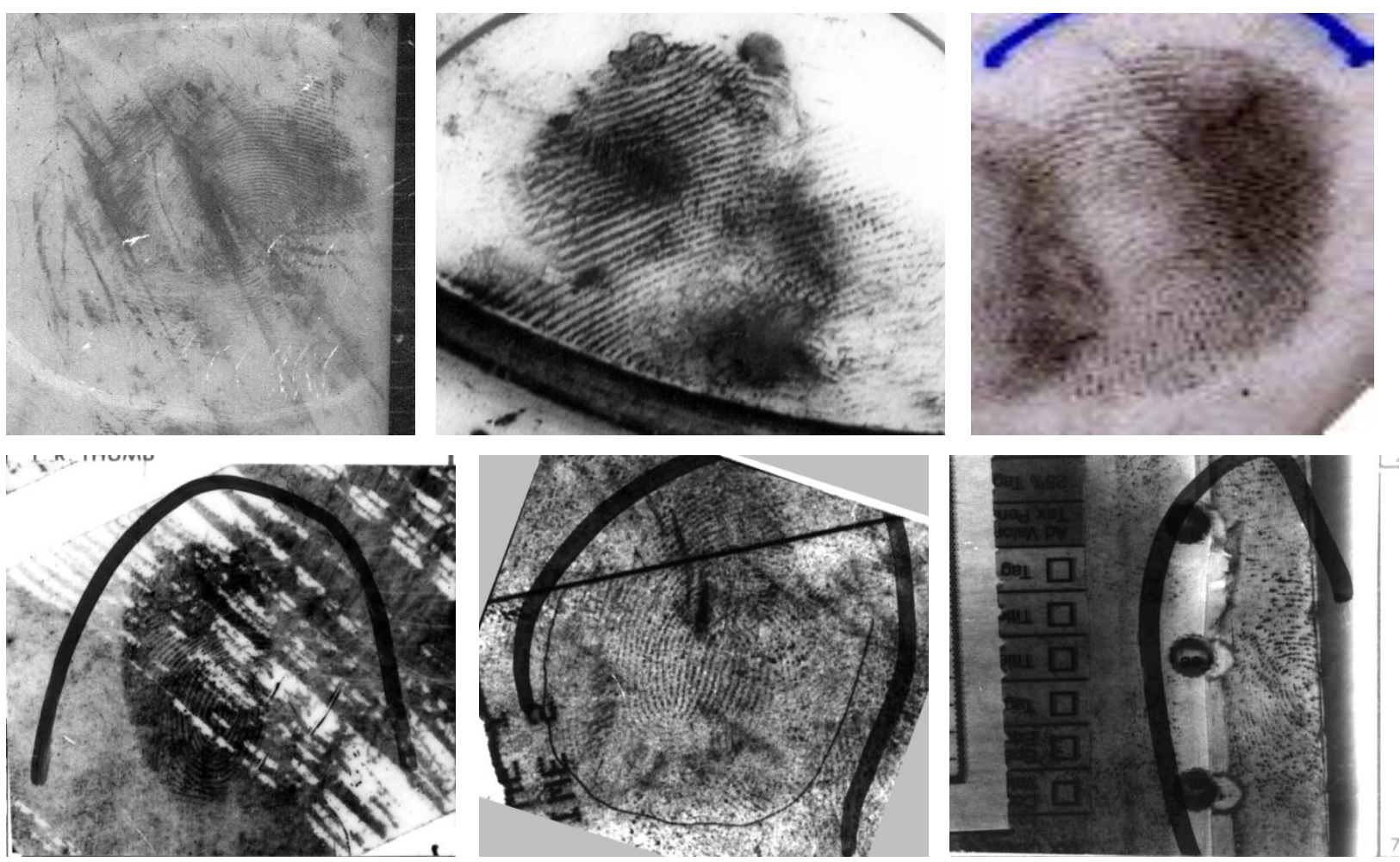}
\caption{Example images from MSP database (top row) and NIST SD27 (bottom row) with RoI markup by a latent examiner (by colored marker).}
\vspace{-1em}
\label{fig:whereLook}
\end{figure}

Latent examiners tend to examine fingermarks, directed by the RoI, to identify the region of interest (see Figure \ref{fig:whereLook}). Thus, by directing attention to a specific fingerprint, we can eliminate unnecessary computation for low interest regions.

We reuse feature maps returned by Faster RCNN to locate the region of interest. Next, we train SegFinNet to learn two classes: (i) attention region (fingermark region identified by a black marker by the examiner (Figure \ref{fig:Full_architecture}) and (ii) fingermark. In the inference phase, a comparison between returned the fingermarks' location to the attention region is used to decide which one needs to be kept using the following criterion: if the overlapping area between the fingerprint bounding box and the attention region is over 70\%, the bounding box is kept.

Our attention mechanism is intuitive, and it helps during matching (see Section \ref{sec:Matching} for more details) because it eliminates background friction ridges which generate spurious minutiae.

\subsection{Feedback scheme}
\label{sec:Feedback_scheme}
One issue in using a detection-based segmentation approach is that it segments objects based on  candidates (RoI returned by detector). Thus, a large proportion of pixels in these bounding boxes belong to the foreground (fingermarks) rather than the background. The need to have a new loss function that can handle the imbalanced class problem is necessary. Let $D=\{(x_i,y_i)\}_{i=1,..,N}$ be a set of $N$ training samples, where $x_i$ is the input image and $y_i$ is its corresponding groundtruth mask. SegFinNet outputs a set of concatenated $C$ masks $\langle (x_i,{\gamma}_{i1}), (x_i,{\gamma}_{i2}), ..., (x_i,{\gamma}_{iC}) \rangle$ for each input $x_i$. We create weights ($w_{j}$, $j=1,..,C$) for each loss value to solve this problem.

\begin{figure}[!tbp]
\centering
\includegraphics[width=8cm]{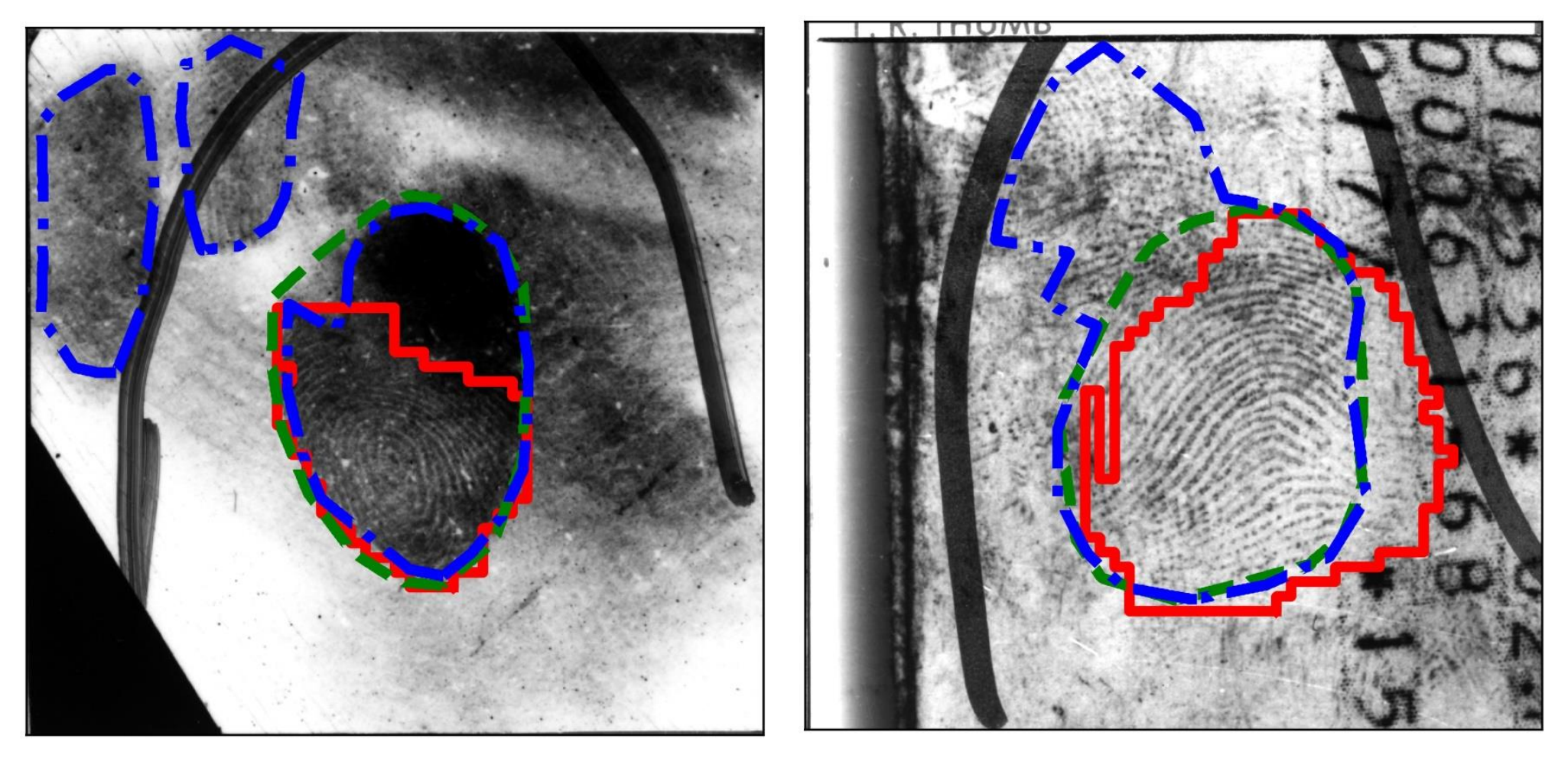}
\caption{Different groundtruths for two latents in NIST SD27. The groundtruth croppings shown in \textcolor{red}{red}, \textcolor{OliveGreen}{green}, and \textcolor{blue}{blue} are used in Cao \etal~\cite{cao2014segmentation}, Ruangsakul \etal~\cite{ruangsakul2015latent}, and Zhu \etal~\cite{zhu2017latent}, respectively.}
\vspace{-1em}
\label{fig:GT_vary}
\end{figure}

Unlike most popular computer vision datasets that have a significant number of pixel-wise annotated masks, there is no dataset available in the fingerprint domain that provides pixel-wise segmentation. Hence, different published studies have used different annotations (see Figure \ref{fig:GT_vary}). Furthermore, since the border of fingermarks is usually not well defined, it leads to inaccuracies in these manual masks and, subsequently, training error.
To alleviate this concern, we propose a semi-supervised partial loss that updates the loss for pixels in the same class while discarding other classes except the background.

Combining the two solutions together, let $\mathcal{L_M}$ be the segmentation (mask) loss which takes into account the proportion of all classes in the dataset:
\vspace{-0.5em}
\begin{equation}
\mathcal{L_M} = \sum_{j \in C} (\sum_{i \in N} w_{j} \mathscr{L}({\gamma}_{ij},y_{ij}) + \lambda l({\gamma}_{ij},y_{ij}))
\label{eq:loss_M}
\end{equation}

where $w_{j}$ is the soft-max weight number of pixels on the $j^{th}$ label, $y_{ij}$ is corresponding mask label of the $i^{th}$ sample in the $j^{th}$ class, $\lambda$ is regularization term, $l(.)$ is cross-entropy loss w.r.t. background, and $\mathscr{L}(.)$ is the per-pixel sigmoid average binary cross-entropy loss as defined in~\cite{he2017mask}.

In the training phase, we consider the loss function as a weight sum of the class, bounding box, and mask loss. Let $\mathcal{L}_{all}, \mathcal{L}_{C}, \mathcal{L}_{B}, \mathcal{L}_{M}$ be the total, class, bounding box, and pixel-wise mask loss, respectively. The total loss for training is calculated as follows:
\vspace{-2mm}
\begin{equation}
\mathcal{L}_{all} = \alpha \mathcal{L_C} + \beta \mathcal{L_B} + \gamma \mathcal{L_M}
\label{eq:loss_SegFinNet}
\end{equation}

We set $\alpha = 2$, $\beta = 1$, $\gamma = 2$ to emphasize the importance of the correctness of predicted class and pixel-wise instance segmentation. We note that the mask loss, $\mathcal{L_M}$ is based on the Intersection over Union (IoU) criteria~\cite{long2015fully, he2017mask, chen2016deeplab}.

\subsection{Voting fusion masks}
\label{sec:Voting_fusion_masks}
\textbf{The effect of grayscale normalization.}
In the computer vision domain, input images usually have intensity values in a specific range. Thus, we can easily recognize, detect or segment objects. However, the fingerprint domain differs from most traditional computer vision related problems. In particular, because of the noisy background and low contrast of the fingerprint ridge structure, mistakes can be easily made in detecting fingermarks. Motivated by the various procedures used by forensic experts to ``preprocess'' images when examining latent fingerprints, we tried different preprocessing techniques on the original latent such as centered gray scaled, histogram equalized, and inverse image.

Even though the original latent fingerprint is noisy, removing noise via an enhancement algorithm prior to segmentation~\cite{cao2014segmentation, zhang2013adaptive} is not advisable because texture information may be lost. 
To make sure the segmentation result is reasonable and invariant to the contrast of input image, we propose a simple but effective voting fusion mask technique. Given an input latent, we first preprocess it to generate different types of grayscale images which are subsequently fed into SegFinNet to get the corresponding score maps. The final score map is accumulated over different grayscale inputs. Each pixel in the image has its own score value. We set the threshold to $K = 3$, which means that each pixel in the chosen region receives at least $3$ votes from the voting masks.

Although this approach seems to increase the computational requirement, it boosts the reliability of the resulting mask while keeping the running time of whole system efficient (see Section \ref{sec:Running_time} for quantitative running time values).

\section{Experiments}
\label{sec:Experiments}

\subsection{Implementation Details}
\label{sec:implement}
We set the anchor size in Faster RCNN varying from $8 \times 8$ to $128 \times 128$. Batch size, detection threshold, and pixel-wise mask threshold are set to $32$, $0.7$, and $0.5$, respectively. The learning rate for SegFinNet is set to 0.001 in the first $30k$ iterations, and 0.0001 in the rest of the $70k$ iterations. Mini-batch size is 1 and weight decay is 0.0001. We set the hyper-parameter $\lambda = 0.8$ in Equation \ref{eq:loss_M} .

\subsection{Datasets}
\label{sec:Dataset}
We have used 3 different latent fingerprint databases: MSP DB (an operational forensic database), NIST SD27~\cite{garris2000nist}, and West Virginia University latent database (WVU)~\cite{WVU}. The MSP DB includes 2K latent images and over 100K reference rolled fingerprints. The NIST SD27 contains $258$ latent images with their true mates while WVU contains $449$ latent images with their mated rolled fingerprints and another $4,290$ non-mated rolled images.

\begin{figure}[!tp]
\centering
\includegraphics[width=8.5cm]{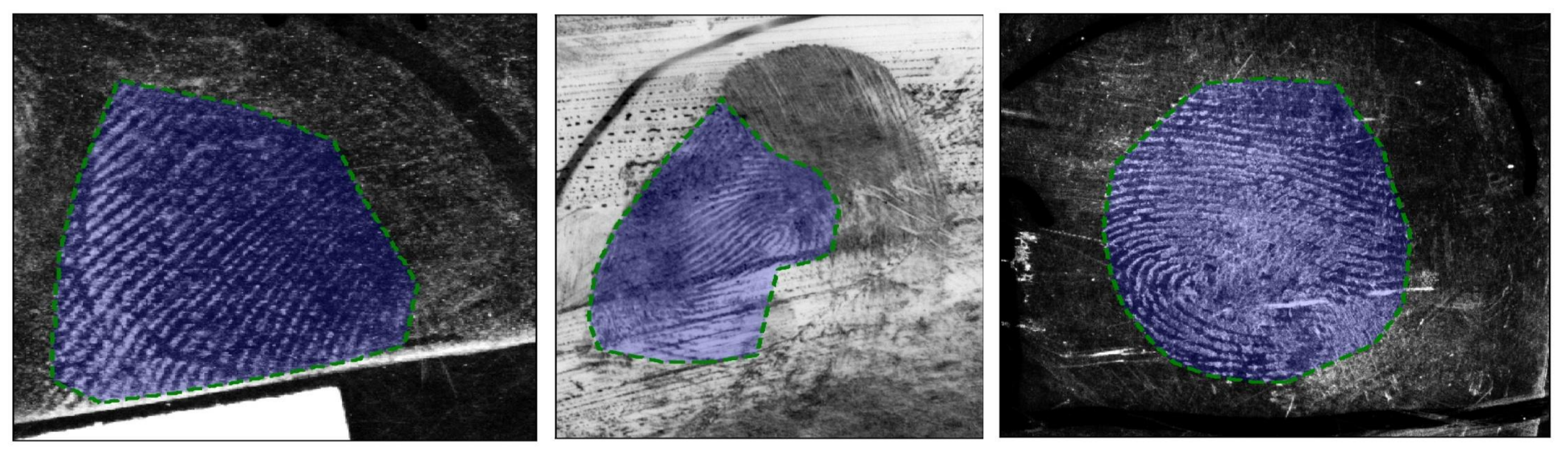}
\caption{Example images in the MSP database with the corresponding manual ground truth mask overlaid.}
\vspace{-1em}
\label{fig:msp}
\end{figure}

\emph{Training:} we used the MSP DB to train SegFinNet. We manually generated ground truth binary masks for each latent. Figure \ref{fig:msp} shows some example latents in the MSU DB with the corresponding groun truth masks. We used a subset of $1000$ latent images from MSP DB for training while using a different set of $1000$ latents in MSP DB for testing. With common augmentation techniques (e.g. random rotation, translation, scaling, cropping, etc. ), the training dataset size increased to $8K$ latent images.

\emph{Testing:} we conducted experiments on NIST SD27, WVU, and 1000 sequestered test images from the MSP database. To make the latent search to appear more realistic, we constructed a gallery consisting of $100K$ rolled images, including the $258$ true mates of NIST SD27, $4,290$ rolled fingerprints in WVU, $27K$ images from NIST14~\cite{NSITDB14}, and the rest from rolled prints in the MSP database. The ground truth masks were obtained from~\cite{jain2011latent}.

\subsection{Cropping Evaluation Criteria}
\label{sec:Criteria}
Published papers based on patch-based approach with a classification scheme reported the cropping performance in terms of MDR and FDR metrics. The lower values of these metrics, the better framework is. Let $A$ and $B$ be two sets of pixels in predicted mask and ground truth mask, respectively. MDR and FDR are then defined as:
\vspace{-0.05em}
\begin{equation}
MDR = \frac{|B| - |A \cap B|}{|B|}   ;
FDR = \frac{|A| - |A \cap B|}{|A|}
\end{equation}

\begin{table}[!bp]
	\centering
	\caption{Comparison of the proposed segmentation method with published algorithms using pixel-wise (MDR, FDR, IoU) metrics on NIST SD27 and WVU latent databases.}
	\vspace{0.7em}
	\label{tab:accuracy}
	\begin{small}
	\begin{threeparttable}
	\begin{adjustbox}{}
		\begin{tabular}{|c|c|c|c|c|}
			\hline
			{\bfseries Dataset}&{\bfseries Algorithm}& {\bfseries MDR}& {\bfseries FDR}& {\bfseries IoU}\\
			\hline
			\hline
			{}&{Choi~\cite{choi2012automatic}\tnote{(\#)}}	&$14.78\%$&$47.99\%$&$43.28\%$\\
			{}&{Zhang~\cite{zhang2013adaptive}}	&	$14.10\%$&	$26.13\%$&$N/A$\\
			{}&{Ruangsakul~\cite{ruangsakul2015latent}\tnote{(\#)}}	&	$24.56\%$&	$36.48\%$&$52.05\%$\\
			{NIST}&{Cao~\cite{cao2014segmentation}\tnote{(\#)}}	&	$12.37\%$&	$46.66\%$&$48.25\%$\\
			{SD27}&{Liu~\cite{liu2016latent}}	&	$13.32\%$&	$24.21\%$&$N/A$\\
			{}&{Zhu~\cite{zhu2017latent}}	&	$10.94\%$& $11.68\%$&$N/A$\\
			{}&{Ezeobiejesi~\cite{ezeobiejesi2017latent}\tnote{(*)}}	&	$1.25\%$&	$0.04\%$&$N/A$\\
			{}&{\bfseries Proposed method}	&	{\bfseries 2.57\%}&	{\bfseries 16.36\%}&{\bfseries 81.76\%}\\
			\hline
			{}&{Choi~\cite{choi2012automatic}}	&$40.88\%$&$5.63\%$&$N/A$\\
			{WVU}&{Ezeobiejesi~\cite{ezeobiejesi2017latent}\tnote{(*)} }	&	$1.64\%$&	$0.60\%$&$N/A$\\
			{}&{\bfseries Proposed method}	&	{\bfseries 13.15\%}&	{\bfseries 5.30\%}&{\bfseries 72.95\%}\\
			\hline
		\end{tabular}
		\end{adjustbox}
		\begin{tablenotes}
		\scriptsize
			\item[(\#)] \scriptsize{We reproduce the results based on masks and ground truth provided by authors.}
			\item[(*)] \scriptsize{Its metrics are on reported patches.}
		\end{tablenotes}
		\end{threeparttable}
	\end{small}
	\vspace{-1em}
\end{table}

With the proposed non-patch-based and top-down approach (detection and segmentation), it is necessary to use IoU metric, which is more appropriate for multi-class segmentation~\cite{long2015fully, he2017mask, chen2016deeplab}. In addition, we report our results in terms of the MDR and FDR metrics for comparison. In contrast to MDR and FDR metrics, a superior framework will lead to a higher value of IoU. The IoU metric is defined as:
\begin{equation}
IoU = \frac{|A \cap B|}{|A \cup B|}
\end{equation}
We note that the published methods~\cite{ruangsakul2015latent, zhu2017latent, choi2012automatic} used their own individual ground truth information so comparing them based on MDR and FDR is not fair given these two metrics critically depend on the ground truth. Figure \ref{fig:GT_vary} demonstrates the variations in ground truths used by existing works\footnote{We contacted the authors to get their groundtruths.}. It is important to emphasize that a favorable metric value does not mean that the associated cropping will lead to better latent recognition accuracy. It simply reveals the overlap between predicted mask and its manually annotated ground truth.

\begin{figure*}[!tbp]
\centering
\includegraphics[width=16.5cm]{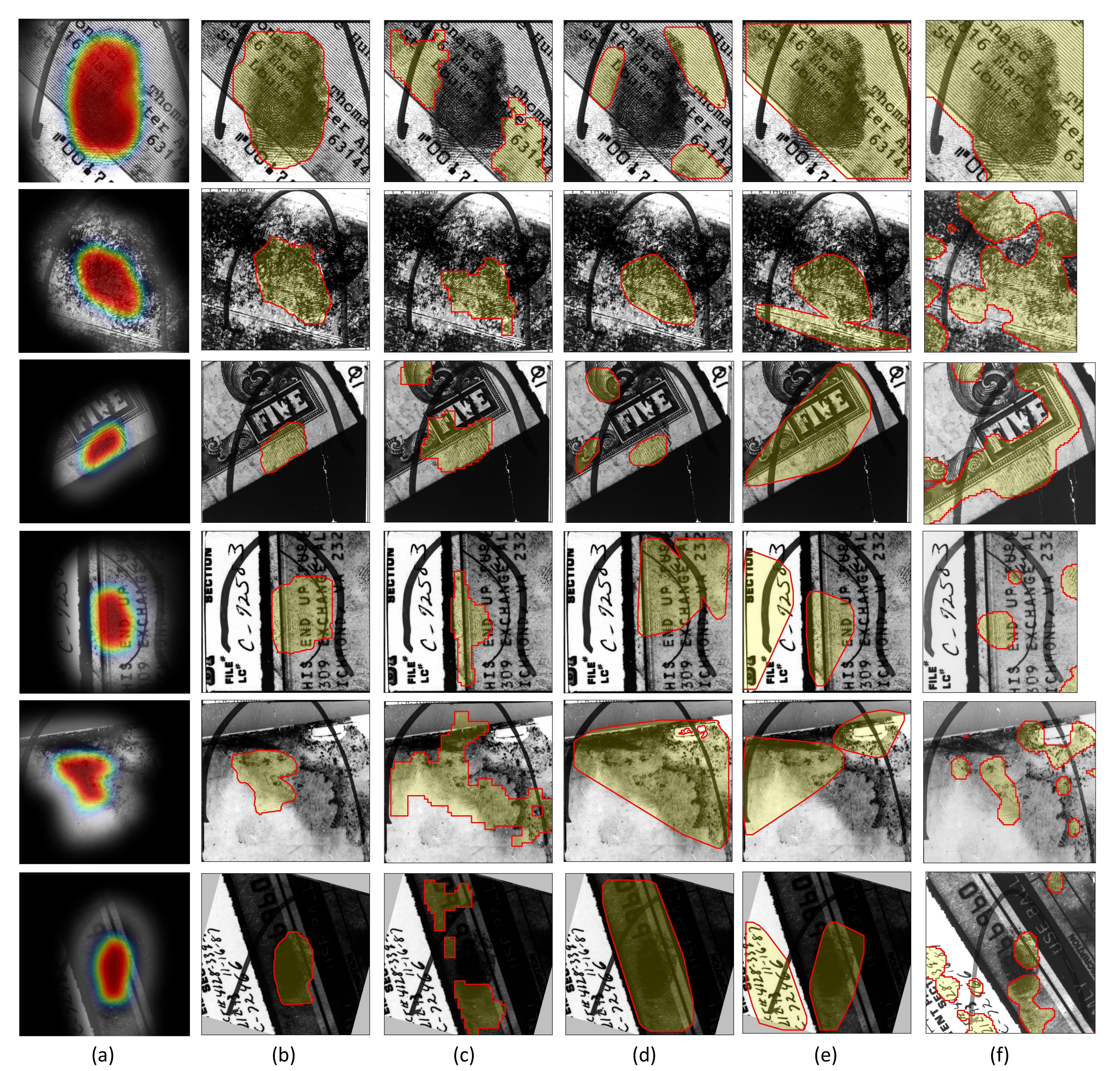}
\caption{Visualizing segmentation results on six different (one per row) latents from NIST SD27. (a) Our visual attention with heatmap (fingermark probability), (b) Proposed method, (c) Ruangsakul \etal~\cite{ruangsakul2015latent}, (d) Choi \etal~\cite{choi2012automatic}, (e) Cao \etal~\cite{cao2014segmentation}, (f) Zhang \etal~\cite{zhang2013adaptive}. Images used for comparison vary in terms of noise, friction ridge area and ridge clarity. Note that Zhang \etal used 1,000 dpi images while others, including us, used 500 dpi latents.}
\vspace{-1em}
\label{fig:vis1}
\end{figure*}

\subsection{Cropping Accuracy}
\label{sec:Accuracy}
Table \ref{tab:accuracy} shows a quantitative comparison between SegFinNet and other existing works using MDR and FDR metrics on NIST SD27 and WVU databases. The IoU metric was computed based on masks and the ground truth provided by the authors. Because Ezeobiejesi \etal evaluated MDR and FDR on patches, it is not fair to include it for IoU comparison. Table \ref{tab:accuracy} also shows that SegFinNet provides the lowest error rate in terms of MDR and FDR. This is because of our use of non-patch based approach. The table also reveals that the low values of either MDR or FDR only does not usually lead to high IoU value.

\subsection{Running time}
\label{sec:Running_time}
Experiments are run on a Nvidia GTX Ti 1080 GPU with 12GB of memory. Table \ref{tab:speed} shows a comparison with different configurations in computation time on NIST SD27 and WVU. We note that the voting fusion technique takes longer time to process an image because it runs on different inputs. However, its accuracy is better than just using a single image with attention technique.

\begin{table}[!tbp]
	\centering
	\caption{Performance of SegFinNet with different configurations. AM: attention mechanism (Section \ref{sec:wheretolook}), VF: voting fusion scheme (Section \ref{sec:Voting_fusion_masks})}    
	\vspace{0.7em}
	\label{tab:speed}
	\begin{small}
	\begin{adjustbox}{max width=\textwidth}
		\begin{tabular}{|c|c|c|c|}
			\hline
			{\bfseries Dataset}&{\bfseries Configuration}& {\bfseries Time(ms)}& {\bfseries IoU}\\
			\hline
			\hline
			{}&{SegFinNet w/o AM \& VF}	&248&46.83\%\\
			{NIST}&{SegFinNet with AM}	&274&50.60\%\\
			{SD27}&{SegFinNet with VF}	&396&78.72\%\\
			{}&{\bfseries SegFinNet full}	&{\bfseries 457}&{\bfseries 81.76\%}\\
			\hline
			{}&{SegFinNet w/o AM \& VF}	&198&51.18\%\\
			{WVU}&{SegFinNet with AM}	&212&62.07\%\\
			{}&{SegFinNet with VF}	&288&67.33\%\\
			{}&{\bfseries SegFinNet full}	&{\bfseries 361}&{\bfseries 72.95\%}\\
			\hline
		\end{tabular}
		\end{adjustbox}
	\end{small}
	\vspace{-1em}
\end{table}

Figure \ref{fig:vis1} shows the visualization of SegFinNet compared to existing works on NIST SD27. These masks were obtained by contacting authors. 

\subsection{Latent Matching}
\label{sec:Matching}
The final goal of segmentation is to increase the latent matching accuracy. We used two different matchers for latent to rolled matching: Verifinger SDK 6.3~\cite{verifinger} and a state-of-the-art latent COTS AFIS. To make a fair comparison to existing works~\cite{zhang2013adaptive, ruangsakul2015latent, liu2016latent}, we also report the matching performance for Verifinger on 27K background from NIST 14~\cite{NSITDB14}. In addition, we report the performance of COTS on 100K background. To explain the  matching experiments, we first define some terminologies.

(a) \emph{Baseline:} Original gray scale latent image.

(b) \emph{Manual GT:} Groundtruth masks from Jain \etal~\cite{cao2014segmentation}.

(c) \emph{SegFinNet with AM:} Masked latent images using visual attention mechanism only.

(d) \emph{SegFinNet with VF:} Masked latent images using majority voting mask technique only.

(e) \emph{SegFinNet full:} Masked latents with full modules.

(f) \emph{Score fusion:} Sum of score fusion of our proposed SegFinNet with SegFinNet+AM, SegFinNet+VF, and original input latent images.

\begin{table}[!tbp]
	\centering
	\caption{Matching results with Verifinger on NIST SD27 and WVU against 27K background.}
	\vspace{0.7em}
	\label{tab:matchVeri}
	\begin{small}
	\begin{adjustbox}{max width=\textwidth}
		\begin{tabular}{|c|c|c|c|}
			\hline
			{\bfseries Dataset}&{\bfseries Methods}& {\bfseries Rank-1}& {\bfseries Rank-5}\\
			\hline
			\hline
			{}&{Choi~\cite{choi2012automatic}\tnote{(\#)}}	&	$11.24\%$&	$12.79\%$\\
			{}&{Ruangsakul~\cite{ruangsakul2015latent}\tnote{(\#)}}	&	$11.24\%$&	$11.24\%$\\
			{NIST}&{Cao~\cite{cao2014segmentation}\tnote{(\#)}}	&	$11.63\%$&	$12.01\%$\\
			{SD27}&{Manual GT}	&	$10.85\%$&	$11.63\%$\\
			{}&{Baseline}	&	$8.14\%$&	$8.52\%$\\
			{}&{\bfseries Proposed method}	&	{\bfseries 12.40\%}&	{\bfseries 13.56\%}\\
			{}&\emph{Score fusion}	&	$13.95\%$&	$16.28\%$\\
			\hline
			{}&{Manual GT}	&	$25.39\%$&	$26.28\%$\\
			{WVU}&{Baseline}	&	$26.28\%$&	$27.61\%$\\
			{}&{\bfseries Proposed method}	&	{\bfseries 28.95\%}&	{\bfseries 30.07\%}\\
			{}&\emph{Score fusion}	&	$29.39\%$&	$30.51\%$\\
			\hline
		\end{tabular}
		\end{adjustbox}
	\end{small}
	\vspace{-1em}
\end{table}

Table \ref{tab:matchVeri} demonstrates matching results using Verifinger. 
Since there are many versions of the Verifinger SDK, we use masks provided by the authors in ~\cite{choi2012automatic, zhang2013adaptive, cao2014segmentation, ruangsakul2015latent} to make a fair comparison. However, the authors did not provide their masks for the WVU database. Note that contrary to popular belief, the manual groundtruth does not always give better results than original images.

\begin{figure}[!tbp]
\centering
\includegraphics[width=8.75cm]{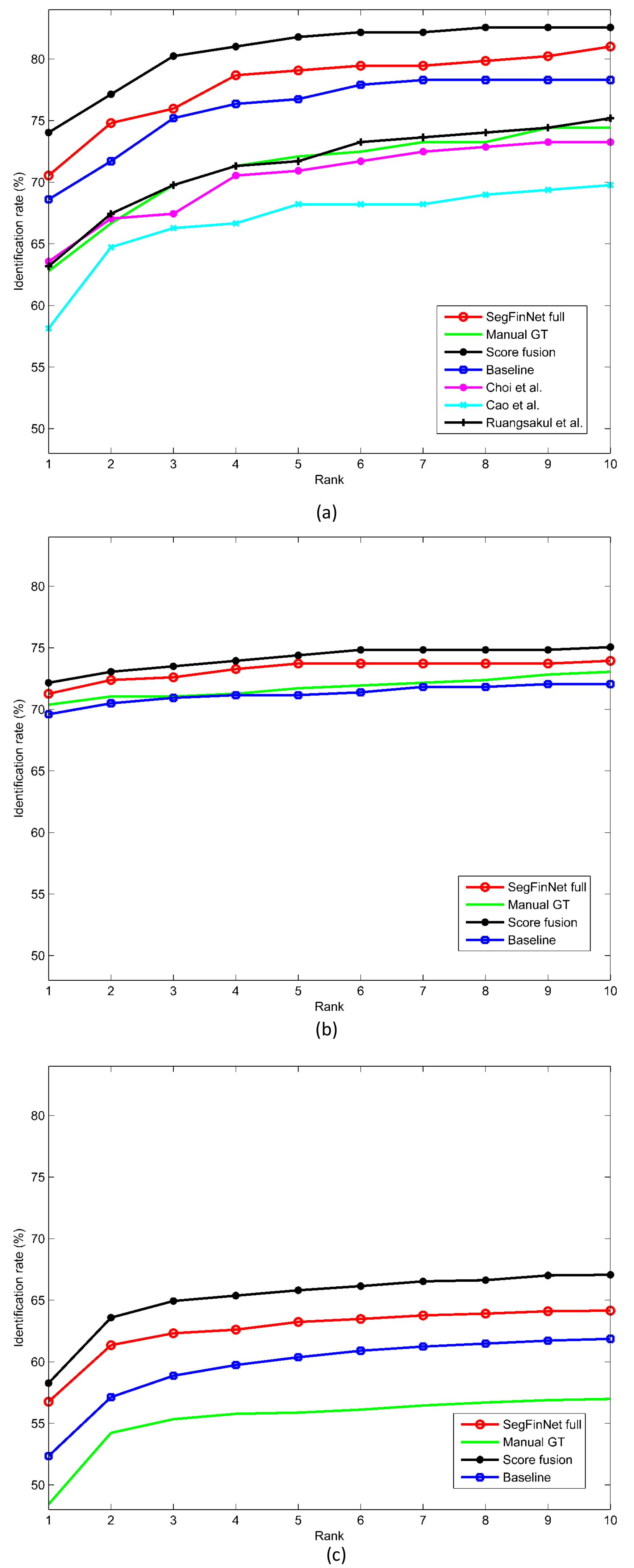}
\caption{Matching results with a state-of-the-art COTS matcher on (a) NIST SD27, (b) WVU, and (c) MSP database against 100K background images.}
\vspace{-1em}
\label{fig:matchCOTS}
\end{figure}

Figure \ref{fig:matchCOTS} shows the matching results using state-of-the-art COTS. We did not use any enhancement techniques like Cao \etal~\cite{cao2014segmentation} in this comparison. The combination between the attention mechanism and the voting technique showed better performance in our proposed method. Besides, highest results of score fusion technique mean that our method can be complementary to using full image in matching.
\vspace{-0.5em}
\section{Conclusion}
\label{sec:Conclusion}
We have proposed a framework for latent segmentation, called SegFinNet. It utilizes a fully convolutional neural network and detection based approach for latent fingerprint segmentation to process the full input image instead of dividing it into patches. Experimental results on three different latent fingerprint databases (i.e. NIST SD27, WVU, and MSP database) show that SegFinNet outperforms both human ground truth cropping for latents and published segmentation algorithms. This improved cropping, in turn, boosts the hit rate of a state of the art COTS latent fingerprint matcher. Our framework can be further developed along the following lines: (a) Integrating into an end-to-end matching model by using shared parameters learned in the Faster RCNN backbone as a feature map for minutiae/non-minutiae extraction; (b) Combining orientation information to get instance segmentation for segmenting overlapping latent fingerprints.

\section*{Acknowledgements}
This research is based upon work supported in part by the Office of the Director of National Intelligence (ODNI), Intelligence Advanced Research Projects Activity (IARPA), via IARPA R\&D Contract No. 2018-18012900001. The views and conclusions contained herein are those of the authors and should not be interpreted as necessarily representing the official policies, either expressed or implied, of ODNI, IARPA, or the U.S. Government. The U.S. Government is authorized to reproduce and distribute reprints for governmental purposes notwithstanding any copyright annotation therein.

{\small
\bibliographystyle{ieee}
\bibliography{BTAS_ref}
}

\end{document}